\PassOptionsToPackage{table, dvipsnames}{xcolor}
\documentclass[lettersize,journal]{IEEEtran}
\usepackage{amsmath,amsfonts}
\usepackage{algorithmic}
\usepackage{algorithm}
\usepackage{array}
\usepackage[caption=false,font=normalsize,labelfont=sf,textfont=sf]{subfig}
\usepackage{textcomp}
\usepackage{stfloats}
\usepackage{url}
\usepackage{verbatim}
\usepackage{graphicx}
\usepackage{cite}
\usepackage[hidelinks]{hyperref}

\usepackage{algorithm}
\usepackage{algorithmic}
\usepackage{xcolor}
\usepackage{booktabs}
\usepackage{multirow}
\usepackage[table]{xcolor}
\usepackage{listings}
\usepackage[dvipsnames]{xcolor}

\usepackage{etoolbox}
\makeatletter
\AfterEndEnvironment{algorithm}{\let\@algcomment\relax}
\AtEndEnvironment{algorithm}{\kern2pt\hrule\relax\vskip3pt\@algcomment}
\let\@algcomment\relax
\newcommand\algcomment[1]{\def\@algcomment{\footnotesize#1}}
\renewcommand\fs@ruled{\def\@fs@cfont{\bfseries}\let\@fs@capt\floatc@ruled
  \def\@fs@pre{\hrule height.8pt depth0pt \kern2pt}%
  \def\@fs@post{}%
  \def\@fs@mid{\kern2pt\hrule\kern2pt}%
  \let\@fs@iftopcapt\iftrue}
\makeatother

\begin{document}

\title{Enhancing Fine-grained Object Detection \\ in Aerial Images via Orthogonal Mapping}

\author{Haoran Zhu, Yifan Zhou, Chang Xu, Ruixiang Zhang, and Wen Yang
\thanks{The research was supported in part by the National Natural Science Foundation of China (NSFC) General Program under Grant 62271355.}
\thanks{H. Zhu, Y. Zhou, C. Xu, R. Zhang, and W. Yang are with the School of Electronic Information, Wuhan University, Wuhan 430072, China. \emph{E-mail: zhuhaoran@whu.edu.cn; zhouyifan75@gmail.com; \{xuchangeis, zhangruixiang, yangwen\}@whu.edu.cn}}

}




\maketitle

\begin{abstract}
Fine-Grained Object Detection (FGOD) is a critical task in high-resolution aerial image analysis. This letter introduces Orthogonal Mapping (OM), a simple yet effective method aimed at addressing the challenge of semantic confusion inherent in FGOD. OM introduces orthogonal constraints in the feature space by decoupling features from the last layer of the classification branch with a class-wise orthogonal vector basis. This effectively mitigates semantic confusion and enhances classification accuracy. Moreover, OM can be seamlessly integrated into mainstream object detectors. Extensive experiments conducted on three FGOD datasets (FAIR1M, ShipRSImageNet, and MAR20) demonstrate the effectiveness and superiority of the proposed approach. Notably, with just one line of code, OM achieves a 4.08\% improvement in mean Average Precision (mAP) over FCOS on the ShipRSImageNet dataset. Codes are released at \href{https://github.com/ZhuHaoranEIS/Orthogonal-FGOD}{https://github.com/ZhuHaoranEIS/Orthogonal-FGOD}.
\end{abstract}


\section{Introduction}
\IEEEPARstart{F}{ine-Grained} Object Detection (FGOD) in aerial images aims to precisely recognize fine-level subcategories while localizing them simultaneously. Unlike generic object detection, which focuses on coarse-level categories like airplanes, ships, or vehicles, FGOD requires identifying specific fine-level subcategories such as C919, ARJ21, and Boeing747 within the category of airplanes. FGOD in aerial images can significantly facilitate the precise interpretation of aerial images~\cite{fair1m}. However, compared to generic object detection, FGOD presents unique challenges due to substantial inter-class variations and intricate intra-class differences, leading to \textbf{semantic confusion}~\cite{finegrained_1}.

Recently, much effort has been put into tackling semantic confusion. These methods can be broadly categorized into two main categories: \textbf{feature representation enhancement} and \textbf{feature space discrimination}. Feature representation enhancement is a classic strategy to tackle semantic confusion since better feature expression capabilities can help fine-grained subcategory classification.
For instance, Sumbul et al.~\cite{multisource} propose a multi-source attention mechanism. At the same time, PETDet~\cite{petdet} enhances proposal quality and feature discrimination capabilities using a quality-oriented proposal network and bilinear channel fusion network. 
Recent works cast new insights into feature space discrimination. 
PCLDet~\cite{pcldet} constructs a prototype bank and prototypical contrastive loss for maximizing inter-class distance and minimizing intra-class distance to extract discriminative features.
Without constructing positive and negative sample pairs or complex additional modules, orthogonal losses~\cite{ole, opl, linearity-enhanced, orthogonalityloss} have been proposed. These methods utilize cosine metrics to enforce orthogonality between features of different classes, thereby increasing inter-class distances.

\begin{figure}[t]
    \centering
    \includegraphics[width=0.99\linewidth]{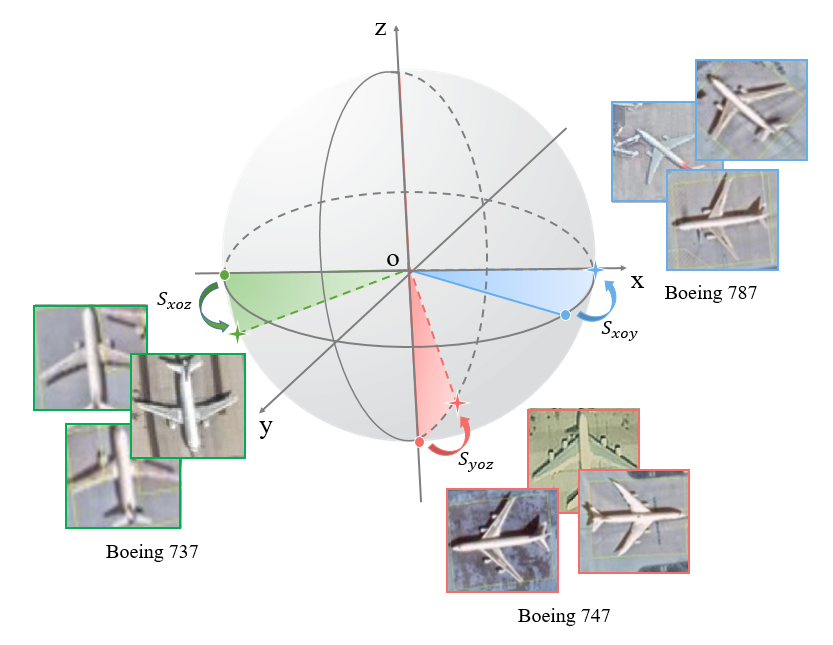}
    \caption{ A toy example of proposed Orthogonal Mapping (OM). Concretely, OM situates distinct fine-grained subcategories in an orthogonal space. As illustrated in the figure, Boeing 737, Boeing 747, and Boeing 787 are positioned in the $S_{xoz}$, $S_{yoz}$, and $S_{xoy}$ spaces, respectively.}
    \label{fig:toyexample}
\end{figure}

Despite these advancements, existing methods still exhibit several limitations. For feature representation enhancement, existing algorithms require additional network parameters or auxiliary supervision to facilitate feature learning, significantly increasing the model’s complexity. For feature space discrimination, contrastive learning-based methods~\cite{contrastive-learning} demand the construction of positive and negative sample pairs, and the effectiveness of these methods is directly influenced by the quality of this process~\cite{similaritycl}. Orthogonal loss-based approaches, while effective in tasks such as face recognition and image classification, do not adequately consider the unique challenges of the FGOD in Aerial Images: weak feature representation in the regions of interest (RoI) and massive negative samples (\textit{i.e.}, background) compared to the image classification task. The weak feature makes orthogonal losses (soft constraints) less effective. The massive negative samples lead to biases towards enforcing orthogonality among the negative samples while neglecting the orthogonality among the positive samples Moreover, these orthogonal losses are typically tailored for modifications based on softmax, making them less compatible with sigmoid-based one-stage algorithms.

To address the above issues, we propose Orthogonal Mapping (OM), a simple yet effective method that can be deployed on both one-stage and two-stage networks to mitigate semantic confusion. Fig.~\ref{fig:toyexample} illustrates a toy example of our proposed method. Specifically, unlike serving as an additional loss function, our OM fully addresses the unique challenges of the FGOD task without introducing additional network parameters. Instead, through hard constraints, our approach assigns completely orthogonal category prototypes to each class and background, thereby mapping different categories to orthogonal spaces and mitigating inter/intra-class confusion.
We integrate OM into the one-stage algorithm FCOS~\cite{fcos} and the advanced two-stage algorithm PETDet~\cite{petdet} and evaluate its performance on three widely recognized FGOD datasets (FAIR1M~\cite{fair1m}, MAR20~\cite{mar20}, and ShipRSImageNet~\cite{shiprsimagenet}), demonstrating its effectiveness.

\section{Methodology}
Prior research \cite{orthogonal} presents the convolution operation as a matrix-vector multiplication, wherein various matrix-vector pairs transform the input tensor into different vector spaces for subsequent learning. In this letter, we aim to enhance inter-class distinctions by promoting independence among these diverse mapping spaces. To achieve this, we manually devise a set of orthogonal vector bases to replace the convolution operation. Firstly, we introduce the perspective of convolution as a matrix-vector multiplication. Next, we delve into the intricacies of the designed orthogonal vector bases. Lastly, we elucidate the seamless integration of these orthogonal vector bases into both the one-stage method FCOS and the two-stage method PETDet.


\subsection{Convolution as a Matrix-Vector Multiplication}
For the last convolutional layer of the classification branch with the input feature tensor $X \in \mathbb{R}^{N \times H \times W}$, where $H$ and $W$ denote the size of the input tensor, and $N$ represents the feature dimension. The convolution kernel $K \in \mathbb{R}^{C \times k \times k \times N}$, where $C$ is the number of classes, and $k$ is the size of the convolutional kernel. The output tensor $Y \in  \mathbb{R}^{H \times W \times C}$. Since the convolution is linear, based on this, we can express the network's output $Y$ as:
\begin{equation}
    Y^{H \times W \times C} = {(K^{C \times k \times k \times N} \cdot X^{N \times H \times W})}^T,
    \label{con:outputY}
\end{equation}
where $\cdot$ denotes matrix multiplication.

\subsection{Orthogonal Vector Bases}
As mentioned earlier, convolution operation can be viewed as matrix multiplication. Therefore, to map the inputs into an orthogonal space, we add orthogonal regularization to the last
layer of the classification branch.
First, we apply AvgPooling on the convolutional kernel $K$ and flatten the convolution $K$ into a two-dimensional matrix, where $K \in \mathbb{R}^{C \times N}$. Then, we employ Gram-Schmidt Orthogonalization along the $C$ dimension to obtain orthogonal vector bases, denoted as $\tilde{K} \in \mathbb{R}^{C \times N}$.
Finally, we rewrite the output as $Y = {\langle \tilde{K}, X \rangle}^T$, where the $\langle \cdot, \cdot \rangle$ operation is defined as follows:
\begin{equation}
    \langle \tilde{K}_{i}, X_{j} \rangle = \frac{\tilde{K}_{i} \cdot X_{j}}{\Vert \tilde{K}_{i} \Vert_2 \cdot \Vert X_{j} \Vert_2}.
    \label{con:cdot}
\end{equation}
where $\Vert \cdot \Vert_2$ represents the $L2$ norm. Thereunder, the output $Y$ can be viewed as the cosine similarity between the input $X$ and each orthogonal space. Following training, $\tilde{K}$ is capable of mapping different categories into distinct orthogonal spaces effectively. It is noteworthy that $\tilde{K}$ remains constant after random initialization.

\begin{algorithm}[t]
\caption{Pseudocode of OM in a PyTorch-like style.}
\label{alg:code}
\algcomment{\fontsize{7.2pt}{0em}\selectfont \texttt{gs}: Gram-Schmidt Orthogonalization; \texttt{cs}: cosine similarity; \texttt{fl}: focal loss.
}
\definecolor{codeblue}{rgb}{0.25,0.5,0.5}
\lstset{
  backgroundcolor=\color{white},
  basicstyle=\fontsize{7.2pt}{7.2pt}\ttfamily\selectfont,
  columns=fullflexible,
  breaklines=true,
  captionpos=b,
  commentstyle=\fontsize{7.2pt}{7.2pt}\color{codeblue},
  keywordstyle=\fontsize{7.2pt}{7.2pt},
}
\begin{lstlisting}[language=python]
# f_k: the convolution kernel (CxNxkxk)
# labels: ground truth labels

f_k.params  # initialize
for x in loader:  # load a minibatch x with B samples
    # contrust orthogonal vector bases
    k = avgpooling(f_k) # k: CxNx1x1
    k = flatten(1) # k: CxN
    k = gs(k) # Orthogonalize matrix k.
    k = k.detach()  # no gradient to the kernel
    
    # output the classification confidence, x: BxHxWxN
    cls_scores = cs(x, k.T)

    # classification loss
    loss = fl(cls_scores, labels)

    # SGD update
    loss.backward()
\end{lstlisting}
\end{algorithm}

\subsection{Application to Object Detectors}
Our approach is designed to be a Plugin and can be compatible with various object detection frameworks. To showcase the adaptability of our Orthogonal Mapping (OM) method, we seamlessly integrate it into two strong baseline models: the one-stage method FCOS and the two-stage method PETDet. 
We replace the last convolutional layer or fully connected layer in the original head with OM to introduce orthogonal constraints.
Specifically, we predefine the orthogonal feature prototypes $v_k$ for each class $k \in \{1, ..., C\}$ (C means the number of categories). For a network's output feature $u$, The probability that the feature $u$ belongs to class $j$ is given by calculating the cosine similarity between $u$ and $v_j$. In this way, features of the same category ($j$) exhibit similar expressions (all lie within the space of $v_j$), while features of different categories ($i,j$) occupy mutually orthogonal spaces ($ v_i \perp v_j$).
Algorithm~\ref{alg:code} provides the pseudo-code of OM.

\begin{table*}[t]
\renewcommand{\arraystretch}{1.0}
\caption{Comparison results on FAIR1M-v1.0 online validation. All the models are with ResNet-50 as the backbone. Moreover, FRCNN, ORCNN, and RoI Trans denote Faster R-CNN, Oriented R-CNN, and RoI Transformer respectively.}
\centering
\begin{tabular}{c c | c c c | c c c c c}
\toprule

\multicolumn{2}{c|}{\multirow{2}{*}{Method}} & FCOS~\cite{fcos}  & S2ANet~\cite{s2anet} & \textbf{FCOS w/ OM} & FRCNN~\cite{faster-rcnn} & ORCNN~\cite{orcnn} & RoI Trans~\cite{roitrans} & PETDet~\cite{petdet} &  \textbf{PETDet w/ OM} \\

 & & \multicolumn{3}{c|}{\textit{one-stage methods}} & \multicolumn{5}{c}{\textit{two-stage methods}} \\

\midrule
\multicolumn{2}{c|}{Backbone} & R-50 & R-50 & R-50 & R-50 & R-50 & R-50 & R-50 & R-50 \\
\midrule
\multirow{10}{*}{Airplane} & B737 & 36.95 & 36.06 & 35.13 & 33.94 & 35.17 & 40.14 & 41.05 & 42.38 \\
 & B747 & 79.70 & 84.33 & 81.78 & 84.25 & 85.17 & 84.92 & 82.23 & 82.70 \\
 & B777 & 12.15 & 15.62 & 12.89 & 16.38 & 14.57 & 15.39 & 22.04 & 20.07 \\
 & B787 & 36.76 & 42.34 & 39.74 & 47.61 & 47.68 & 49.17 & 51.21 & 55.71 \\
 & C919 & 1.48  & 1.95  & 11.64 & 14.44 & 11.68 & 19.73 & 25.66  & 22.91 \\
 & A220 & 48.48 & 44.09 & 46.72 & 47.40 & 46.55 & 50.46 & 51.83 & 53.61 \\
 & A321 & 66.04 & 68.00 & 66.52 & 68.82 & 68.18 & 70.31 & 69.53 & 70.11 \\
 & A330 & 61.19 & 63.84 & 61.19 & 72.71 & 68.60 & 71.42 & 70.86 & 71.16 \\
 & A350 & 70.62 & 70.00 & 70.14 & 76.53 & 70.21 & 72.62 & 73.70 & 71.46 \\
 & ARJ21& 4.14  & 12.10 & 26.84 & 26.59 & 25.32 & 33.65 & 36.90 & 30.60 \\
 \midrule
\multirow{8}{*}{Ship} & PS & 8.54 & 8.82  & 8.95 & 11.03 & 15.20 & 13.21 & 13.29  & 14.56 \\
 & MB  & 52.52 & 48.03 & 53.66 & 51.22 & 60.42 & 56.54 & 61.94 & 63.36 \\
 & FB  & 8.565  & 6.79  & 8.82 & 6.41  & 9.10  & 6.82  & 8.90  & 9.30  \\
 & TB  & 30.52 & 34.01 & 33.98 & 34.19 & 36.83 & 35.71 & 37.88 & 37.63 \\
 & ES  & 11.87 & 7.49  & 10.93 & 9.41  & 11.32 & 9.96  & 10.93 & 11.64 \\
 & LCS & 13.82 & 18.30 & 13.70 & 15.17 & 21.86 & 16.91 & 22.05 & 23.36 \\
 & DCS & 37.39 & 37.62 & 36.52 & 32.26 & 38.22 & 36.01 & 36.82 & 36.82 \\
 & WS  & 23.00 & 22.96 & 22.21 & 11.27 & 22.67 & 17.30 & 23.98 & 24.05 \\
 \midrule

\multirow{9}{*}{Vehicle} & SC & 52.22 & 61.57 & 52.13 & 54.56 & 57.62 & 58.29 & 68.81 & 68.64 \\
 & BUS & 22.00 & 11.76 & 21.51 & 22.94 & 24.40 & 28.02 & 18.33 & 30.02 \\
 & CT  & 33.20 & 34.28 & 34.44 & 37.74 & 40.84 & 40.55 & 42.17 & 42.76 \\
 & DT  & 28.00 & 36.03 & 28.91 & 41.69 & 45.20 & 45.97 & 47.15 & 47.63 \\
 & VAN & 45.93 & 54.62 & 46.94 & 48.23 & 54.01 & 54.10 & 65.48 & 65.36 \\
 & TRI & 11.11 & 3.47  & 9.73  & 12.46 & 15.46 & 11.82 & 11.93  & 10.73 \\
 & TRC & 2.45  & 0.96  & 1.21  & 2.44  & 2.37  & 2.61  & 2.09   & 2.08  \\
 & EX  & 11.21 & 7.24  & 12.12 & 11.35 & 13.55 & 11.74 & 8.51  & 10.58 \\
 & TT  & 0.44  & 0.40  & 0.53  & 0.32  & 0.24  & 0.72  & 0.41   & 0.55  \\
 \midrule

\multirow{4}{*}{Court} & BC & 40.97 & 38.44 & 42.18 & 45.18 & 48.18 & 47.50 & 42.64 & 45.35 \\
 & TC & 78.41 & 80.44 & 78.17 & 77.75 & 78.45 & 80.12 & 78.60 & 79.54 \\
 & FF & 53.69 & 56.34 & 57.24 & 52.05 & 60.79 & 58.03 & 63.51 & 63.10 \\
 & BF & 87.56 & 87.47 & 86.93& 87.19 & 88.43 & 87.37 & 87.40 & 87.21 \\
 \midrule

\multirow{3}{*}{Road} & IS & 55.79 & 50.76 & 56.13 & 58.71 & 57.90 & 58.58 & 57.48 & 57.11 \\
 & RA & 17.39 & 16.67 & 19.20 & 19.38 & 17.57 & 21.85 & 20.82 & 24.86 \\
 & BR & 15.08 & 17.24 & 12.53 & 20.76 & 28.63 & 23.52 & 22.57 & 23.44 \\
 \midrule

\multicolumn{2}{c|}{AP$_{0.5}$} & 34.08 & 34.71 & \textbf{35.33} & 36.83 & 38.85 & 39.15 & 40.55 & \textbf{41.19} \\
\bottomrule
\end{tabular}
\label{table:fair1mv1detail}
\end{table*}

\section{Experiments}
\subsection{Datasets}
To verify our method on the FGOD task, we conduct the main experiments and visual analysis on FAIR1M-v1.0 dataset~\cite{fair1m}. The testing results of FAIR1M-v1.0 are
submitted for validation through the official evaluation server. We also verify the effectiveness of the proposed method on two additional fine-grained aerial image datasets, MAR20~\cite{mar20} and ShipRSImageNet~\cite{shiprsimagenet}.

\subsection{Implementation Details}
We build the code using MMrotate~\cite{mmrotate} based on the PyTorch~\cite{pytorch} deep learning framework. All models are trained on one single NVIDIA GeForce RTX3090 GPU with a batch size of 8. We use the Stochastic Gradient Descent (SGD) optimizer for 12 epochs with 0.9 momentum and 0.0001 weight decay for models trained on the FAIR1M dataset. The initial learning rate is set to 0.005 and decays at the $8^{th}$ and ${11}^{th}$ epochs. All the other parameters are set the same as default in MMrotate. As for MAR20 and ShipRSImageNet, models are trained with 36 epochs, and the learning rate is reduced at the factor of 0.1 in the ${24}^{th}$ and ${33}^{rd}$ epochs. The preprocessing and data augmentation strategies for all datasets follow PETDet~\cite{petdet}.

\begin{figure*}[t]
    \centering
    \includegraphics[width=0.85\linewidth]{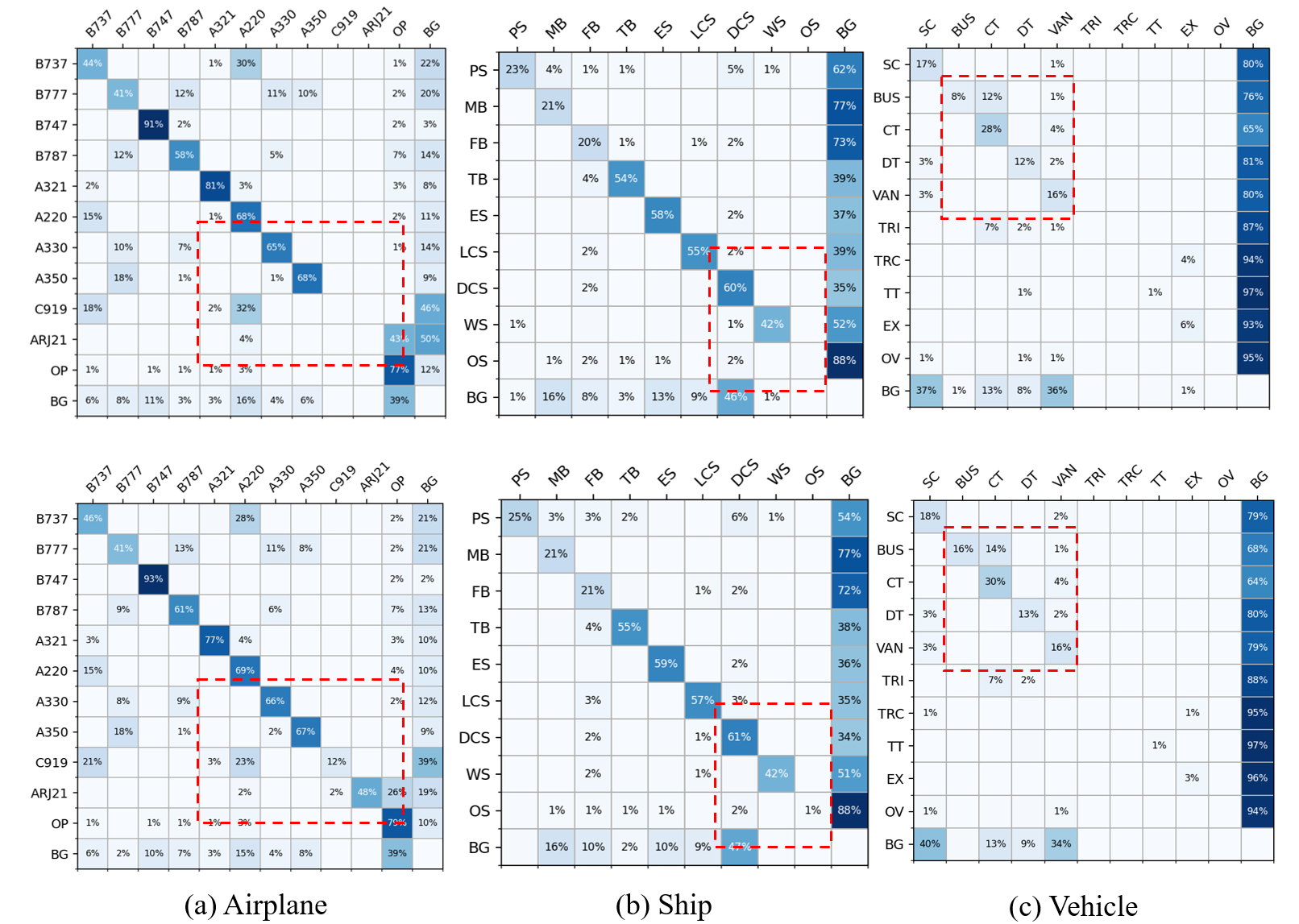}
    \caption{ Confusion matrices of detection results (\%) obtained from FCOS w/o OM (top) and FCOS w/ OM (bottom). The horizontal and vertical coordinates represent the ground truth labels and the prediction labels. (a) Airplane. (b) Ship. (c) Vehicle.}
    \label{fig:confusionmatrix}
\end{figure*}

\begin{table}[t]
\renewcommand{\arraystretch}{1.5}
	\centering
	\caption{Comparison results between our OM and other SOTA orthogonal losses on the ShipRSImageNet.}
	\resizebox{1.0\linewidth}{!}{
    \begin{tabular}{l | l | c | c | c }
    \toprule
    Stages & Method & $\rm{mAP}$(\%) & Param.(M) & FLOPs(G) \\
    \midrule
    \multirow{4}{*}{One-stage Methods} & FCOS~\cite{fcos} & 26.58 & 31.97 & 127.08 \\
    \cline{2-5}
    & FCOS w/ OLÉ~\cite{ole} & 24.36 & 31.97 & 127.08 \\
    \cline{2-5}
    & FCOS w/ OPL~\cite{opl} & 26.19 & 31.97 & 127.08 \\
    \cline{2-5}
    & FCOS w/ OM & \textbf{30.66} & \textbf{31.89} & \textbf{125.95} \\
    \hline
    \hline
    \multirow{4}{*}{Two-stage Methods} & PETDet~\cite{petdet} & 47.74 & 47.67 & 130.08 \\
    \cline{2-5}
    & PETDet w/ OLÉ~\cite{ole} & 47.64 & 47.67 & 130.08 \\
    \cline{2-5}
    & PETDet w/ OPL~\cite{opl} & 48.13 & 47.67 & 130.08 \\
    \cline{2-5}
    & PETDet w/ OM & \textbf{49.00} & \textbf{47.63} & \textbf{130.04} \\
    \bottomrule
\end{tabular}
\label{tab:shiprs}}
\end{table}


\begin{table}[t]
\renewcommand{\arraystretch}{1.3}
\caption{Comparison results on MAR20.}
\centering
\begin{tabular}{c | c  c c}
\toprule
Method & mAP & AP$_{0.5}$ & AP$_{0.75}$\\
\midrule
FCOS~\cite{fcos} & 52.94 & 78.90 & 62.72\\
FCOS w/ OM & 53.85$^{+0.91}$ & 79.52$^{+0.62}$ & 63.88$^{+1.16}$\\
\midrule
PETDet~\cite{petdet} & 59.85 & 84.85 & 75.78 \\
PETDet w/ OM & 60.04$^{+0.19}$ & 84.66$^{-0.19}$ & 75.90$^{+0.12}$ \\
\bottomrule
\end{tabular}
\label{table:shiprsmar}
\end{table}

\subsection{Main Results}
\textbf{FAIR1M:} We conduct the main experiments on the FAIR1M-v1.0 datasets. Table \ref{table:fair1mv1detail} presents the comparative results of our proposed Orthogonal Mapping (OM) and other methods on FAIR1M-v1.0. 
By categorizing studies into one-stage and two-stage methods, we have the following observations. First, incorporating OM into both the one-stage method FCOS and the two-stage method PETDet yields measurable improvements, validating the effectiveness of our approach. Second, the improvements are relatively modest when OM is integrated into PETDet. This is attributed to PETDet's already substantial improvements in addressing semantic confusion. Additionally, our method focuses on enhancing feature discrimination through orthogonal characteristics without improving feature representation, thus limiting the extent of the improvements. Third, certain categories, such as the Tractor in the Vehicle class, show less substantial improvement after integrating our method. 
This is mainly attributed to its limited presence in the dataset (only 0.07\% of the total dataset), while our OM does not inherently enhance the feature representation capability of individual categories, hence the limited improvement.

\textbf{Other Datasets:} To further validate the effectiveness of our proposed method, we conduct experiments on two additional fine-grained aerial image datasets: ShipRSImageNet and MAR20. The results are summarized in Table \ref{tab:shiprs} and \ref{table:shiprsmar}. For FCOS, integrating the proposed OM yields a substantial performance improvement. However, the enhancement achieved by applying OM to PETDet is not as pronounced. As previously discussed, one-stage methods typically encounter more severe issues with semantic confusion compared to two-stage methods. In our experiments, we observe that PETDet already attains high accuracy across various categories in the MAR20 and ShipRSImageNet datasets, suggesting that semantic confusion is less prevalent. Consequently, the incremental improvement introduced by our method is marginal.

\textbf{Comparison with orthogonal losses:} We compared our method with the SOTA orthogonal losses OPL and OLÉ on the ShipRSImageNet, as shown in Table~\ref{tab:shiprs}. In terms of accuracy, OM outperforms the other algorithms. Notably, incorporating OPL and OLÉ into the FCOS results in negative impacts. This is because OPL and OLÉ are enhancements based on the softmax function, while FCOS uses a sigmoid-based focal loss, leading to adverse effects. In the PETDet, our approach also surpasses the other two. This is due to our method employing a hard constraint, which does not require additional learning objectives. In contrast, OPL and OLÉ, as soft constraints, necessitate extra learning tasks (\textit{i.e.}, ensuring orthogonality in the feature space), which are not well-suited for the inherently weaker features in FGOD compared to the image classification.
In terms of computational burden and network parameters, OPL and OLÉ, being solely loss functions, do not introduce additional network parameters but do increase the computational load. Our method, on the other hand, replaces the original convolutional layer, thereby reducing both parameter count and computational complexity.

\subsection{Visual Analysis}
\begin{figure}[t]
    \centering
    \includegraphics[width=0.99\linewidth]{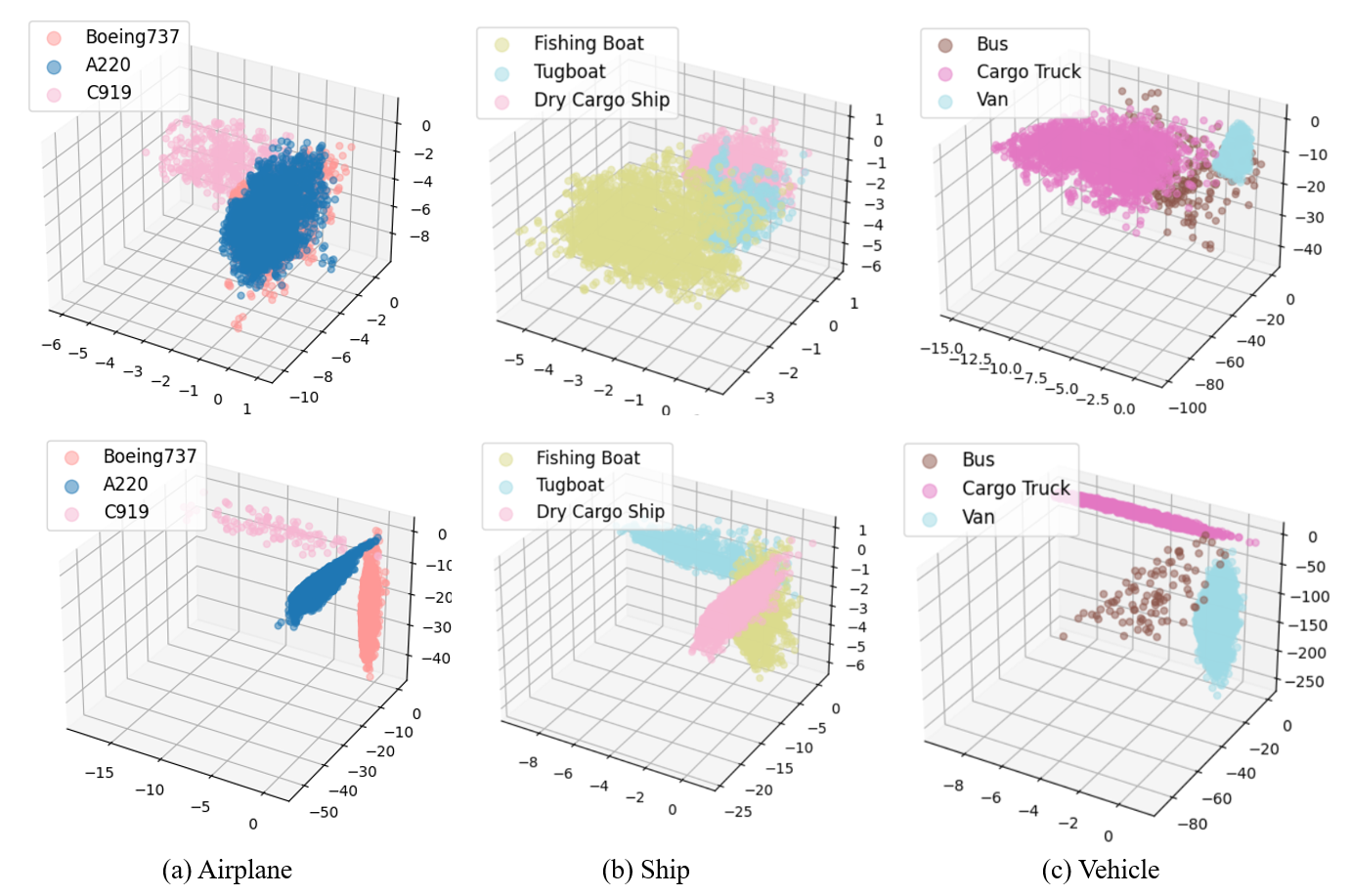}
    \caption{
    The three-dimensional distribution of three main easily confused classes in FAIR1M-v1.0 obtained from FCOS w/o OM (top) and FCOS w/ OM (bottom). (a) Airplane. (b) Ship. (c) Vehicle.}
    \label{fig:featuremap}
\end{figure}

We conduct a series of visualization experiments to validate our method's effectiveness in mapping different categories to an orthogonal space to mitigate semantic confusion. First, we visualize the confusion matrix on the FAIR1M-v1.0 dataset, as shown in Fig. \ref{fig:confusionmatrix}. The top row represents the baseline method, while the bottom row represents the results after applying our OM. The visualization demonstrates that our method effectively alleviates the issue of semantic confusion. Particularly noteworthy is the significant improvement observed in categories such as C919 and ARJ21 following the application of our method. 
Second, we present three-dimensional feature distribution plots for subcategories within the Airplane, Ship, and Vehicle classes, known for their semantic confusion. To clearly illustrate the orthogonal effect, we limit the three-dimensional visualization to three categories. The results, depicted in Fig. \ref{fig:featuremap}, show that upon integrating OM, the feature spaces of these categories become orthogonal to each other. This indicates that our method effectively enhances the discrimination between these categories, thus mitigating semantic confusion.


\section{Conclusion}
Fine-grained object detection presents a notable challenge due to semantic confusion, wherein distinguishing between closely related categories becomes difficult. This letter introduces Orthogonal Mapping (OM), a simple yet effective method designed to mitigate semantic confusion in fine-grained object detection. By mapping different categories to orthogonal spaces, OM effectively enhances the discriminative capacity of object detectors. Importantly, our method can be seamlessly integrated into existing object detection frameworks with minimal modification, requiring just a single line of code replacement. Experiments on three fine-grained aerial image datasets validate the effectiveness and superiority of our approach.




\end{document}